\documentclass[letterpaper, 10 pt, conference]{ieeeconf}  % Comment this line out
\IEEEoverridecommandlockouts                              % This command is only
\usepackage[utf8]{inputenc}
\usepackage[T1]{fontenc}
\usepackage{mathtools}
\usepackage[table]{xcolor}
\usepackage{array}
\newcolumntype{P}[1]{>{\centering\arraybackslash}p{#1}}

\title{\LARGE \bf
A Deep Learning Forecaster with Exogenous Variables for Day-Ahead Locational Marginal Price
}

\author{Dipanwita Saha$^{1}$ and Felipe Lopez$^{1}$% <-this % stops a space
\thanks{*This work was supported by GE Digital's Operations Performance Management program.}% <-this % stops a space
\thanks{$^{1}$D. Saha and F. Lopez are with GE Digital, Austin, 
 Texas, USA.
        {\tt\small \{dipanwita.saha, luis.lopez2\}@ge.com}}
}

\begin{document}

\maketitle
\thispagestyle{empty}
\pagestyle{empty}

%%%%%%%%%%%%%%%%%%%%%%%%%%%%%%%%%%%%%%%%%%%%%%%%%%%%%%%%%%%%%%%%%%%%%%%%%%%%%%%%
\begin{abstract}

Several approaches have been proposed to forecast day-ahead locational marginal price (daLMP) in deregulated energy markets. The rise of deep learning has motivated its use in energy price forecasts but most deep learning approaches fail to accommodate for exogenous variables, which have significant influence in the peaks and valleys of the daLMP. Accurate forecasts of the daLMP valleys are of crucial importance for power generators since one of the most important decisions they face is whether to sell power at a loss to prevent incurring in shutdown and start-up costs, or to bid at production cost and face the risk of shutting down. In this article we propose a deep learning model that incorporates both the history of daLMP and the effect of exogenous variables (e.g., forecasted load, weather data). A numerical study at the PJM independent system operator (ISO) illustrates how the proposed model outperforms traditional time series techniques while supporting risk-based analysis of shutdown decisions.

\end{abstract}

%%%%%%%%%%%%%%%%%%%%%%%%%%%%%%%%%%%%%%%%%%%%%%%%%%%%%%%%%%%%%%%%%%%%%%%%%%%%%%%%
\section{INTRODUCTION}

Electricity is a special commodity. Electricity is economically non-storable, which creates the need for constant balancing of locational supply and demand. The independent system operator (ISO), entity responsible for maintaining the system and managing services, balances demand with supply from power generators in two energy cash markets: (1) Day-ahead, which transacts for generation of energy on the next day, with every hour being transacted separately; and (2) Real-time, a reconciliation market that clears any deviations from predicted schedules and ensures balance.

This article covers only the day-ahead market. In the day-ahead market, generators use their best knowledge about production cost (calculated with an estimate of the natural gas price) and estimated power demand (published by ISOs) to construct bids for future energy deliveries. Generators participate by submitting 24 separate bids (one per hour). However, even when the auction is conducted independently for each hour, the generation costs across the day are not independent due to the presence of start-up costs. Since some units (e.g., combined-cycle power plants) require a costly start-up and a time lag until they become online, generators must adjust their bid curve to ensure that they will be allowed to produce energy during continuous periods, even if they include hours with negative spark spread~\cite{wen2001strategic}.

If next day prices were known with certainty, the generator could adjust (lower) their bids to be allowed to operate even in hours with negative spark spread as long as the total loss is less than shutdown and start up costs. The challenge is that next day prices are never known with certainty. Power prices are affected by changes in demand (e.g., weather effects and the intensity of business and daily activities) and supply (e,g., competitors' scheduled outages), which can be drastic and make power prices highly volatile~\cite{weron2014electricity}.

Significant effort has been dedicated to forecast energy price with approaches that include fundamental models, multi-agent models, statistical models, and machine learning models. Refer to Weron~\cite{weron2014electricity} for a detailed review. In this article we take on the challenge of developing a forecaster for day-ahead locational marginal price (daLMP) that can support risk analysis to determine when it is more profitable to produce at a loss than to shut down.

We propose a deep learning model that estimates daLMP to enable risk-based analysis of bidding decisions. The rest of the paper is organized as follows. Section \ref{sec:background} reviews the state of the art on machine learning forecasters for daLMP. Section \ref{sec:DL} describes both the predictors that were selected for use in the neural network, and the network architecture that combines endogenous and exogenous signals. Section \ref{sec:benchmark} shows numerical results that compare our proposed model with traditional time series models on data from the PJM ISO. Section \ref{sec:analysis} provides a discussion of the numerical results. Section \ref{sec:risk} illustrates the use of the proposed model in risk management. Finally, section \ref{sec:future} presents ideas for future work and section \ref{sec:conclusions} concludes the paper.

\section{BACKGROUND}
\label{sec:background}

Forecasting daLMP is a problem that has attracted the attention of power generators since the rise of market deregulation in the 1990s. Power generators have dedicated significant effort to the design of optimal bidding policies that would allow them to maximize their profitability. Often the design of an optimal bidding policy depends on energy prices. Power generation is profitable as long as
\begin{equation}
C(\hat{P}-HR \times \hat{G}) \ge SC \label{eq:profit}
\end{equation}
\noindent where $C$ is unit capacity, $HR$ is unit heat rate, $SC$ is start-up cost, $\hat{P}$ is power price, and $\hat{G}$ is natural gas price. With certain knowledge of day-ahead power prices, generators could adjust their bids so that they clear during the periods when power generation is profitable, even if $\hat{P} < HR \times \hat{G}$~\cite{weron2014electricity}.

The challenge is the calculation of accurate daLMP estimates. The non-storability of electricity and its susceptibility to weather, human activities, and competitor's behavior makes this problem unique. The signal is known to have seasonality at the daily, weekly, and annual levels, and abrupt, short-lived price spikes. The existence of trend, seasonality, and exogenous factors made daLMP ideal for traditional time series analysis techniques. Time series models are attractive to power generators because of their interpretability (physical interpretation may be attached to every factor) but are often criticized for their limited ability to model nonlinear behavior (e.g., effect of ambient temperature in energy consumption)~\cite{weron2014electricity}.

The standard time series model to account for the random nature and time correlations is the \emph{autoregressive moving average} (ARMA) model. Due to the susceptibility of daLMP to shocks from exogenous variables, the standard models are often extended to \emph{autoregressive moving average with exogenous inputs} (ARMAX) models. Standard time series models can be extended to integrated models (ARIMAX), seasonal models (SARIMAX), and heteroscedastic models (GARCH). Some of the earliest references can be tracked to the work of Nogales, Contreras, Conejo, and Esp\'inola~\cite{nogales2002forecasting,contreras2003arima,nogales2006electricity}, who explored various time series approaches to predict daLMP in Spain and California using demand as an exogenous variable. Weron and Misiorek~\cite{weron2005forecasting,misiorek2006point} revisited the problem and reported that the use of exogenous variables and nonlinear factors improves the predictive power of the models. J\'onsson \emph{et al.}~\cite{jonsson2012forecasting} and Cruz \emph{et al.}~\cite{cruz2011effect} highlighted the effect of renewable energies in day-ahead energy prices.

Various extensions were proposed to the time series forecasting models. Some notable contributions are wavelet-ARIMA and neural network hybrid approach~\cite{pindoriya2008adaptive,shafie2011price}; and Haldrup and Nielsen~\cite{haldrup2006regime}, who identified a strong support for long memory and fractional integration. Knittel and Roberts~\cite{knittel2005financial} analyzed the statistical properties of time series models and reported a high degree of persistence and an ``inverse leverage effect'', where positive shocks to price result in large increases in volatility. Zareipour \emph{et al.}~\cite{zareipour2006application} reported that various publicly available sources improve the accuracy of predictive models in ``normal'' operating conditions but unusually high and low prices are still unpredictable.

Machine learning methods, especially neural networks, have also been used to predict daLMP. Although they have been observed to be better suited to capture nonlinear dependencies with predictors, machine learning models have not been widely adopted in energy markets due to their lack of interpretability, which prevents engineers and ISOs from understanding the market's behavior. Some uses of multilayer perceptron neural networks were reported by Cruz \emph{et al.}~\cite{cruz2011effect} and Chen \emph{et al.}~\cite{chen2012electricity}.

A known issue with multilayer perceptron networks is their lack of memory (i.e., predictions are independent of previous network state). As an alternative, recurrent neural networks (RNNs) represent dynamical systems where the network's state can depend both on the network inputs and on the network's previous state. Anbazhagan and Kumarappan~\cite{anbazhagan2014day} showed that RNNs can outperform traditional forecasting models.

Forgetting factors are often added to allow networks to learn long-term behavior without the \emph{vanishing gradient} problem. A solution that overcomes the vanishing gradient problem with the use of forgetting behavior is the \emph{nonlinear autoregressive model with exogenous inputs} (NARX) model proposed by Lin \emph{et al.}~\cite{lin1996learning}. Surprisingly, ARX models were not used for daLMP forecasting until recently, with the work Andalib and Atry~\cite{andalib2009multi}, and Cha{\^a}bane~\cite{chaabane2014hybrid}.

The recent rise of deep learning has motivated a revisit of daLMP forecasting methods. \emph{Long short-term memory} (LSTM) networks are RNNs where each unit is composed of an input gate, an output gate and a forget gate that regulate the flow of information between units. Qin \emph{et al.}~\cite{qing2018hourly} have shown that LSTMs are suitable for forecasting daLMP. Polson and Sokolov~\cite{polson2020deep} demonstrate the use of deep learning with extreme value theory to predict peak loads in the grid.

When using daLMP estimates to guide bidding policies (the objective in this article), model interpretability is not a big a concern as model accuracy and that tilts the interest in favor of machine learning models. A limitation of some machine learning models is the lack of exogenous variables, which have a significant effect on the valleys of the daLMP signal. Accurate estimation of the daLMP valleys is critical because shutting down when the energy price is low may prevent the generator from losses, but an incorrectly scheduled shutdown can conduct to higher losses in missed revenue.

\section{DEEP LEARNING MODEL}
\label{sec:DL}

\subsection{Exogenous variables} \label{exogenous}

The effect of exogenous variables in daLMP is well known. Power generators can use forecasted regional loads provided by the ISO (e.g., PJM Data Miner~\cite{PJMdataminer}) and weather data from weather forecasters (e.g., weather.com~\cite{weatherAPI}) as predictors in daLMP models. Our model will use only regional loads and temperatures at large cities in the ISO regions as exogenous variables. In the case of the PJM ISO, we use the exogenous variables shown in Table \ref{tb:exogenous}. Forecasted regional loads are used in addition to forecasted aggregate demand to account for network congestion. Readers may include other predictors (e.g., natural gas price forecasts, renewable energy production) as exogenous variables in their own implementations.

\begin{table}[ht]
\renewcommand{\arraystretch}{1.3}
\centering
\caption{Exogenous variables used in the PJM region}
\begin{tabular}{l c}
\hline
\textbf{Predictor}          & \textbf{Unit} \\ \hline
Aggregate demand RTO        & MW            \\
Demand in AEP zone          & MW            \\
Demand in APS zone          & MW            \\
Demand in DOM zone          & MW            \\
Demand in MIDATL zone       & MW            \\
Demand in EKPC zone         & MW            \\
Demand in ATSI zone         & MW            \\
Demand in COMED zone        & MW            \\
Demand in DUQ zone          & MW            \\
Temperature in Chicago      & deg F         \\
Temperature in Cincinnati   & deg F         \\
Temperature in Philadelphia & deg F         \\
Temperature in Pittsburgh   & deg F        \\ \hline
\end{tabular}
\label{tb:exogenous}
\end{table}

\subsection{Probability distribution of conditional daLMP}
\label{log daLMP}
Energy price is known to be susceptible to large spikes, which may cause lasting distortions in a time series model. As an alternative, a logarithmic transformation can be used to attenuate the effect of large spikes in energy price ~\cite{weron2004modeling}. In this article, we model daLMP following a log-normal distribution. In other words, we train a neural network to predict the \emph{natural logarithm of daLMP} for each hour of the following day.

\subsection{Network architecture}
%Motivation
\subsubsection{Motivation} \label{motivation}

Let us model daLMP as the time series $\{Y_{t-j} \}_j$, with time delays $j = \{0, 1, \ldots, k \}$. Given the sequence $\{Y_{t-j} \}_j$, a traditional LSTM can be trained to predict an n-period ahead forecast into the future, namely, $\{\hat{Y} _{t+i}  \}_i$, for $i=\{1,2, \ldots, n\}$ . In doing so, the LSTM learns the historical trends and predicts the future based only on its past history. However, if there are some real-time events that affect the forecast (i.e., exogenous variables), a traditional LSTM cannot incorporate that effect in its training process. 

Suppose at each future time step ${t+i}$, for $i= 1,2,\ldots,n$, there are some exogenous features ${X_{t+i}}$ with values or forecasts that are already available (e.g., weather forecast, load forecast, time of the day, day of the week) that may affect the behavior of $\{{Y}_{t+i}\}$. In this article, we propose an LSTM network that aims to forecast $\{{Y} _{t+i}\}$ based on its historical trends $\{Y _{t-j}\}_{j=1}^k$ and exogenous variables $\{X_{t+i}\}_{i=1}^n$. Consequently, this network models a nonlinear function F such that $
\{\hat{Y} _{t+i}\}_{i=1}^n = F(\{Y _{t-j}\}_{j=1}^k , \{X _{t+i}\}_{i=1}^n) $.
 
%Architecture
\subsubsection{Architecture} \label{architecture}
Following the motivation in section \ref{motivation} the model is split into two blocks: (1) a recurrent block that can learn and model the temporal trend using historical daLMP prices at times $\{t-k,\ldots, t\}$, and (2) a parallel CNN block that can extract features from the exogenous side  at times $t+i$. Subsequently, the output of the two blocks are stacked to learn a causal filter which extracts the effect that the exogenous inputs have on the forecasted LMP at time $t+i$ \emph{independently} of any other time point. In this paper, we propose the above network to predict a 24-period ahead forecast of the daLMP prices.

The inputs for the model, as described above, are divided into two categories; the endogenous inputs and the exogenous inputs corresponding to the recurrent and CNN block, respectively. Endogenous inputs consist of past sequential logarithmic daLMP as described in section \ref{log daLMP} and exogenous inputs, which represents direct information of a future events having no temporal dependency, includes regional loads of PJM, weather information as described in section \ref{exogenous} along with one hot encoded variables like time of the day and day of the week.  Using the distinction, the inputs of the are
\begin{itemize}
\item Endogenous Inputs: $Z =\{Z_S^q\}_{q=1}^{m} \in {\rm I\!R}^{m\times n_z\times 1}$,  where $m$ is the batch size and $n_z$ is the LSTM time step per training example.
\item Exogenous Inputs: $X = \{X_F^q\}_{q=1}^{m}\in {\rm I\!R}^{m \times n_x \times x_F}$, where $n_x$ is the length of forecast window into the future, and  $x_F$ is the number of exogenous features per training example. 
\end{itemize}

\begin{figure*}[ht]
\centering
\includegraphics[width=0.95\textwidth]{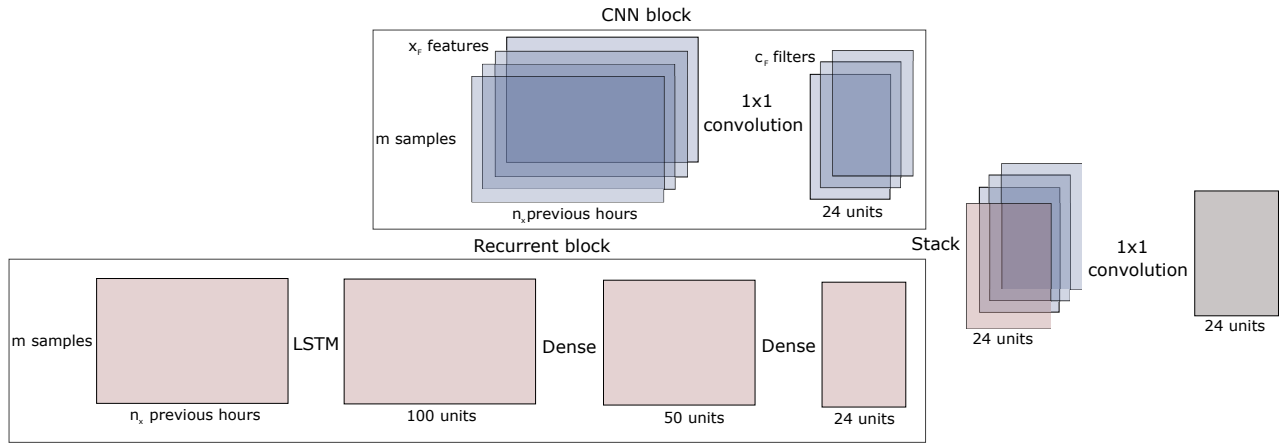}
\caption{Network architecture showing the contribution of historical data (recurrent block) and exogenous variables (CNN block) on daLMP.}
\label{fig:prob_LSTM}
\end{figure*}

Note that, ($\{Z_S^q\}_{q=1}^{m}$,$\{X_S^q\}_{q=1}^{m}$)  is the $q^{th}$ training example pair, such that $\{Z_S^q\}=\{Z_{t}^q,Z_{t-1}^q,\ldots,Z_{t-n_y}^q\}$ is a vector of fixed length representing past information and $\{X_F^q\} = \{X_{t+1}^q,X_{t+2}^q,\ldots,X_{t+n_x}^q\}$ is the corresponding exogenous features where $X_k^q \in {\rm I\!R}^{x_F}$.
During training phase, the network trains two parallel blocks, the recurrent and the CNN block. The recurrent block ingest the $m$ sequences ${Z}_{q=1}^{n_y }$ and passes it through a LSTM layer having 100 neurons followed by two consecutive dense layers having 50 and 24 neurons respectively. Thus, the recurrent block outputs a 3D forecasted daLMP, based on past trend, of dimension $m \times 24 \times 1$. Simultaneously, the CNN block  takes the corresponding 3D  exogenous feature set $X$  and uses a 1D convolution and $c_F$ filters to extract intermediate features for each time in the $n_x$ forecast window. The output dimension of the CNN block is $m \times 24 \times x_F$. Subsequently, the output of these two blocks are concatenated into one tensor having dimension $m \times 24 \times y_F+c_F$. Let us call this layer $L_{concat}$. The output layer,  $L_{out}$ is obtained by passing the concatenated tensor in $L_{concat}$ through a 1D convolution layer with 1 filter to obtain a $m \times 24 \times 1$ tensor. Note that the 1D convolution on the concatenated tensor ensures that the effect of each exogenous feature at time $t$ is cascaded down to the daLMP forecast, obtained from the recurrent block, at that same time without being perturbed by the influence of exogenous variable at any other times in the forecast window. The architecture is shown in Figure \ref{fig:prob_LSTM}.

In the prediction phase, the test examples $(Z,X)$ are passed one at a time, i.e, $m$ is 1 and the model outputs a vector that contains the natural logarithm of daLMP.

The proposed architecture is trained by minimizing the mean absolute error (MAE). In particular, given the training set $\{(Z_k,X_k), y_k\}_{k=1}^m$, where $y_k = \{y_k^1, y_k^2,\ldots, y_k^{24}\}$, the vector of 24 hour ahead predictions of the kth training example, we minimize $\frac{1}{m}\sum_{k=1}^m\left\Vert\log(y_k) - F(X_k,Z_k,w)\right\Vert_1$, where $w$ is the set of the network weights and F is the network map that maps the  exogenous $(X_k)$ and endogenous $(Z_k)$ input to the logarithm of 24-hour ahead prediction of daLMP ($\log(y_k)$). The mean absolute error (MAE) was chosen over the  traditional mean square error (MSE) to mitigate the impact of shocks in the signals. MSE uses Euclidean norm which would put more weight on the high prices which will result in training a biased model. The Adam optimizer was used to solve the above optimization problem.

%Hyperparameters
\subsubsection{Hyperparameters}
\begin{itemize}
\item Activation Function: All layers within the network use the same activation function. The common activation function used is the ReLU activation  as day-ahead price can never be negative.
\item Early stopping parameters: Different regularization techniques including dropout and early stopping were investigated. However, dropout showed a detrimental effect on sequential models, therefore early stopping was used instead.The early stopping parameters to tune are min$\_$delta: which denotes the minimum change in the validation loss to qualify as an improvement and
patience: the number of epochs with no improvement after which training is discontinued. 
\item LSTM sequence length ($n_z$): For the LSTM structure, each input is modeled as a sequence of past values. Considering that values too far in the past do not cause any effect in the day-ahead prices, selecting the right length for the input sequences might remove unnecessary complexities like exploding/vanishing gradient. Therefore, a third hyper parameter is used to select the length of the input sequences on the endogenous side.
\item CNN Filter size ($c_F$): The size of the filter in the CNN block to learn intermediate exogenous features.
\item Batch size ($m$): The number of samples processed before the network weights are updated.
\end{itemize}

\subsection{Deep Learning Forecaster configuration}
\label{sec:model_config}

In this section we define the network configuration and the hyperparameters used to train the Deep Learning(DL) forecaster described in Section \ref{architecture}. The optimal structure of the recurrent block consists of a LSTM layer with 100 neurons followed by two dense layers with 50 and 24 neurons respectively. The sequence length $n_z$ to the recurrent block is 10 days of past data.  The CNN block uses a 1D convolution with a filter size $(c_F)$ of 3 on the exogenous inputs. The batch size $(m)$ for batch training was fixed to 50 and ReLU activation is used for all layers. The maximum epoch for training the model is capped at 250 and validation based early stopping was used to avoid overfitting.The validation set used $7\%$ of the total data and the early stopping parameters were set to min$\_$delta=0.0001 and patience=20. Additionally, none of the layers require dropout.

\section{Benchmark models for electricity price forecasting}
\label{sec:benchmark}

The proposed deep learning forecaster in Section \ref{sec:model_config} was compared with three other forecasters:
\begin{itemize}
    \item \emph{Model 1}: Time series without exogenous variables
    \item \emph{Model 2}: Time series with exogenous variables
    \item \emph{Model 3}: Stateless neural network model
\end{itemize}

Models 1 and 2 are used to illustrate the improvement of a nonlinear method with respect to more traditional time series techniques, with and without exogenous variables. Model 1 is an AR(6) time series model while model 2 is a SARIMAX (6,0,0)x(2,0,0,24) model. The orders of models 1 and 2 were selected based on the Bayesian Information Coefficient. Model 3 is a stateless neural network model where daLMP is assumed to depend only on exogenous variables (demand forecasts), i.e., model 3 assumes that supply in the electricity market is inelastic.

\begin{figure*}[ht] \label{fig:Results}
\centering
\includegraphics[width=0.97\textwidth]{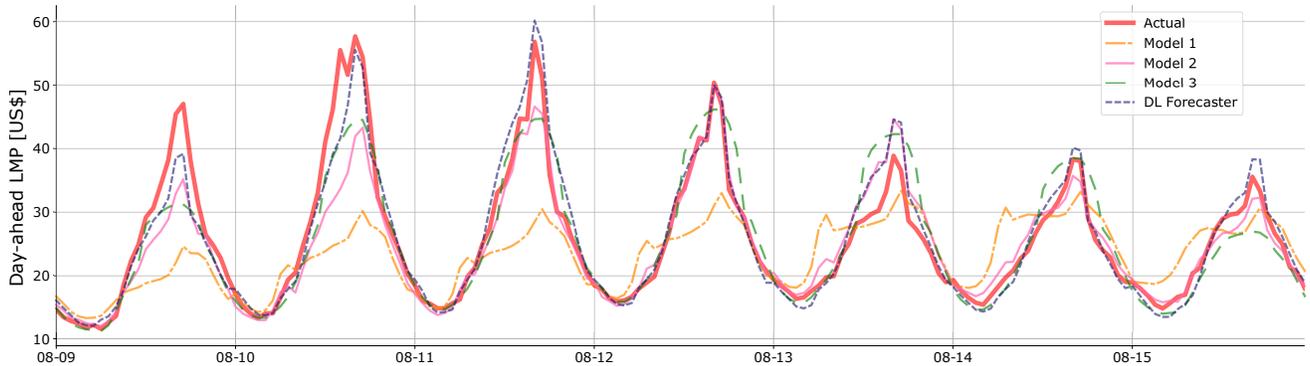}
\caption{Comparison of daLMP forecasts for the week between August 9, 2020 and August 15, 2020 for a node in the PJM network. Each forecast was made with all the information available the day before. Models 1, 2, and 3 were retrained every day with new available information; while the proposed deep learning model (DL Forecaster) was not retrained.}
\label{fig:results}
\end{figure*}

\section{NUMERICAL ANALYSIS}
\label{sec:analysis} 

\subsection{ Data}
In this section, we perform an numerical study to evaluate and analyze the predictive accuracy of the DL forecaster with the benchmark models described in \ref{sec:benchmark}. We the conduct the numerical study for the day-ahead market for a node in PJM ISO. The model was trained on data over a period of 1 year from August 09, 2019 to August 08, 2020. The frequency of each input variables as mentioned in \ref{exogenous} and \ref{log daLMP} is 1 hour, thus, the input variable set consists of $365 \times 24 = 8760$ data points for each input. We compared the predictive accuracy of the four models using the metrics discussed in section \ref{metrics} on one week of out-of-sample data from August 09, 2020 to August 15, 2020. Note that, since there are 24 prices corresponding to the 24 hours of a day, each model predicts 24 daLMP prices, thus, the out-of-sample forecast consists of $7 \times 24= 168$ forecasts. The results are presented in section \ref{results}.

\subsection{Performance Metrics}
\label{metrics}
We compare the three base models proposed in section \ref{sec:benchmark} along with the DL Forecaster \ref{sec:model_config} on 2 metrics, the mean squared error (MSE) and mean absolute percentage error (MAPE). 
The MSE and MAPE calculates the squared deviation of the forecast from the actual and MAPE is the mean percentage of absolute deviation from the forecasted value respectively. 
Let $\hat{p}_k= [\hat{p}_1,\hat{p}_2, \ldots \hat{p}_n]$ be the forecast to the actual daLMP value $p_k= [p_1,p_2, \ldots p_n]$ then,

\begin{equation*}
MSE = \frac{1}{n}\sum_{k=1}^{n}\left\Vert p_{k} - \hat{p}_{k}\right\Vert_2 
\end{equation*}
\begin{equation*}
MAPE = \frac{1}{n}\sum_{k=1}^{n}\left\Vert\frac{p_{k} - \hat{p}_{k}}{p_{k}}\right\Vert_1 \times 100 \%
\end{equation*}
\noindent where $n$ is the the total number of forecasted points. 

\subsection{ Results}
\label{results}
From table \ref{tab:results} we see that the deep learning (DL) forecaster outperforms the other base models on both the MSE and MAPE metrics. Model 1 is unable to follow the peaks and valleys of the daLMP signal, which illustrates the importance of exogeneous variables in daLMP forecasting. Model 3 has an adequate performance (minor issues on the peaks) and that shows that the inelastic supply assumption upon which the model is based is valid in the test set. Model 2 and the DL forecaster outperform Models 1 and 3, with the DL forecaster showing superior performance on both MSE and MAPE.

\begin{table}[ht]
\renewcommand{\arraystretch}{1.3}
\caption{Statistical metrics for daLMP forecasts}
\label{tab:results}
\centering
\begin{tabular}{c|c c}
\hline
{} & \textbf{MSE} &  \textbf{MAPE (\%)}  \\
\hline
Model 1 & 69.17 & 18.0  \\
Model 2 & 17.01 & 8.0  \\
Model 3 & 19.42 & 10.0  \\
Deep learning forecaster & 8.40 & 7.0  \\ \hline
\end{tabular}
\end{table}

\section{RISK-AWARE BIDDING}
\label{sec:risk}

Forecasted daLMP can be used to estimate the risk of incorrectly shutting down when the energy price was higher than the fuel cost and the risk of running when the fuel cost is higher than the energy price. The deep learning model from Section \ref{sec:DL} can be extended with an auxiliary model that represents the error as in the $\log(\mbox{daLMP})$ forecasts as independent zero-mean Gaussian variables, i.e., $\epsilon_i \sim \mathcal{N}(0, \sigma^2)$\footnote{Alternatively, the proposed deep learning model can be expressed as a probabilistic deep learning model with the use of a probabilistic layer (e.g., Tensorflow probability library) or with an uncertainty-aware deep learning models~\cite{lakshminarayanan2017simple}.}. The distribution of profit can be calculated with
\begin{equation}
    p(\mbox{Profit} | HR, \hat{G}) = C \left( \mbox{daLMP} - HR \times \hat{G} \right) - SC
    \label{eq:prob_profit}
\end{equation}
\noindent where daLMP and its distribution are provided by the deep learning model.

Equation \ref{eq:prob_profit} can be used to recommend a shutdown based on the probability that such a decision will be more profitable than its alternative. For instance, shutdown could be recommended when $p(\mbox{Profit} < 0) > 0.90$, i.e., when there is 90\% confidence that a shutdown will be result in a smaller loss than running the unit. Note that a high confidence value is required to make a shutdown decision since the opportunity of cost of an incorrect shutdown is usually larger than the savings from shutting down when daLMP is low.

\section{FUTURE WORK}
\label{sec:future}

The deep learning model and the shutdown policy presented in this article can be extended with separate models that forecast future values for heat rate (known to vary with weather conditions) and natural gas price. The daLMP forecaster can also be extended to automatically optimize hyperparameters (see \ref{sec:model_config}), to include uncertainty estimates, and to remove the log-normal distribution assumption, as mentioned in Section \ref{sec:risk}. Another idea would be to extend the current study to automatically generate optimal bid curves to submit to the ISO using an optimal bidding policy or an adaptive strategy (e.g., reinforcement learning).

\section{CONCLUSIONS}
\label{sec:conclusions}

In this paper we introduce a deep learning architecture that can be adapted to accommodate both historical data and the effect of exogenous variables for predicting day-ahead electricity prices. We present the results of the DL forecasters with traditional time series models like ARMAX, SARIMAX as well as a stateless neural network model as described in section \ref{sec:benchmark}. Note that, the SARIMAX model and the ARMAX allow for incorporating the effect of exogenous variable with historical look back but fail to learn any nonlinear behavior (e.g., effect of temperature in energy consumption or relationship between forecasted regional loads). On the other hand, the stateless neural network depends only on exogenous variable and does not consider any temporal dependency on historical daLMP prices. Therefore, it is unable to capture the underlying trend for future forecast. The deep learning forecaster is a unified solution that eliminate all the drawbacks mentioned above. Following the results in table \ref{tab:results}, we conclude that the daLMP forecasts from the deep learning forecaster outperforms traditional forecasters in common statistical metrics (MSE, MAPE), and is able to capture the peaks and the valleys as compared to the other models.


%%%%%%%%%%%%%%%%%%%%%%%%%%%%%%%%%%%%%%%%%%%%%%%%%%%%%%%%%%%%%%%%%%%%%%%%%%%%%%%%

\section*{ACKNOWLEDGMENT}

The authors thank Yan Liu for his recommendations and for the enlightening discussions on energy markets and bidding policies.

\bibliography{main}

\begin{thebibliography}{10}
\providecommand{\url}[1]{#1}
\csname url@samestyle\endcsname
\providecommand{\newblock}{\relax}
\providecommand{\bibinfo}[2]{#2}
\providecommand{\BIBentrySTDinterwordspacing}{\spaceskip=0pt\relax}
\providecommand{\BIBentryALTinterwordstretchfactor}{4}
\providecommand{\BIBentryALTinterwordspacing}{\spaceskip=\fontdimen2\font plus
\BIBentryALTinterwordstretchfactor\fontdimen3\font minus
  \fontdimen4\font\relax}
\providecommand{\BIBforeignlanguage}[2]{{%
\expandafter\ifx\csname l@#1\endcsname\relax
\typeout{** WARNING: IEEEtran.bst: No hyphenation pattern has been}%
\typeout{** loaded for the language `#1'. Using the pattern for}%
\typeout{** the default language instead.}%
\else
\language=\csname l@#1\endcsname
\fi
#2}}
\providecommand{\BIBdecl}{\relax}
\BIBdecl

\bibitem{wen2001strategic}
F.~Wen and A.~David, ``Strategic bidding for electricity supply in a day-ahead
  energy market,'' \emph{Electric Power Systems Research}, vol.~59, no.~3, pp.
  197--206, 2001.

\bibitem{weron2014electricity}
R.~Weron, ``Electricity price forecasting: A review of the state-of-the-art
  with a look into the future,'' \emph{International Journal of Forecasting},
  vol.~30, no.~4, pp. 1030--1081, 2014.

\bibitem{nogales2002forecasting}
F.~J. Nogales, J.~Contreras, A.~J. Conejo, and R.~Esp{\'\i}nola, ``Forecasting
  next-day electricity prices by time series models,'' \emph{IEEE Transactions
  on Power Systems}, vol.~17, no.~2, pp. 342--348, 2002.

\bibitem{contreras2003arima}
J.~Contreras, R.~Espinola, F.~J. Nogales, and A.~J. Conejo, ``{ARIMA} models to
  predict next-day electricity prices,'' \emph{IEEE Transactions on Power
  Systems}, vol.~18, no.~3, pp. 1014--1020, 2003.

\bibitem{nogales2006electricity}
F.~J. Nogales and A.~J. Conejo, ``Electricity price forecasting through
  transfer function models,'' \emph{Journal of the Operational Research
  Society}, vol.~57, no.~4, pp. 350--356, 2006.

\bibitem{weron2005forecasting}
R.~Weron, A.~Misiorek \emph{et~al.}, ``Forecasting spot electricity prices with
  time series models,'' in \emph{Proceedings of the European Electricity Market
  EEM-05 Conference}, 2005, pp. 133--141.

\bibitem{misiorek2006point}
A.~Misiorek, S.~Trueck, and R.~Weron, ``Point and interval forecasting of spot
  electricity prices: Linear vs. non-linear time series models,'' \emph{Studies
  in Nonlinear Dynamics \& Econometrics}, vol.~10, no.~3, 2006.

\bibitem{jonsson2012forecasting}
T.~J{\'o}nsson, P.~Pinson, H.~A. Nielsen, H.~Madsen, and T.~S. Nielsen,
  ``Forecasting electricity spot prices accounting for wind power
  predictions,'' \emph{IEEE Transactions on Sustainable Energy}, vol.~4, no.~1,
  pp. 210--218, 2012.

\bibitem{cruz2011effect}
A.~Cruz, A.~Mu{\~n}oz, J.~L. Zamora, and R.~Esp{\'\i}nola, ``The effect of wind
  generation and weekday on {Spanish} electricity spot price forecasting,''
  \emph{Electric Power Systems Research}, vol.~81, no.~10, pp. 1924--1935,
  2011.

\bibitem{pindoriya2008adaptive}
N.~Pindoriya, S.~Singh, and S.~Singh, ``An adaptive wavelet neural
  network-based energy price forecasting in electricity markets,'' \emph{IEEE
  Transactions on Power Systems}, vol.~23, no.~3, pp. 1423--1432, 2008.

\bibitem{shafie2011price}
M.~Shafie-Khah, M.~P. Moghaddam, and M.~Sheikh-El-Eslami, ``Price forecasting
  of day-ahead electricity markets using a hybrid forecast method,''
  \emph{Energy Conversion and Management}, vol.~52, no.~5, pp. 2165--2169,
  2011.

\bibitem{haldrup2006regime}
N.~Haldrup and M.~{\O}. Nielsen, ``A regime switching long memory model for
  electricity prices,'' \emph{Journal of Econometrics}, vol. 135, no. 1-2, pp.
  349--376, 2006.

\bibitem{knittel2005financial}
C.~R. Knittel and M.~Roberts, ``Financial models of deregulated electricity
  prices: An application to the {California} market,'' \emph{Energy Economics},
  vol.~27, no.~5, pp. 791--817, 2005.

\bibitem{zareipour2006application}
H.~Zareipour, C.~A. Ca{\~n}izares, K.~Bhattacharya, and J.~Thomson,
  ``Application of public-domain market information to forecast ontario's
  wholesale electricity prices,'' \emph{IEEE Transactions on Power Systems},
  vol.~21, no.~4, pp. 1707--1717, 2006.

\bibitem{chen2012electricity}
X.~Chen, Z.~Y. Dong, K.~Meng, Y.~Xu, K.~P. Wong, and H.~Ngan, ``Electricity
  price forecasting with extreme learning machine and bootstrapping,''
  \emph{IEEE Transactions on Power Systems}, vol.~27, no.~4, pp. 2055--2062,
  2012.

\bibitem{anbazhagan2014day}
S.~Anbazhagan and N.~Kumarappan, ``Day-ahead deregulated electricity market
  price forecasting using neural network input featured by {DCT},''
  \emph{Energy Conversion and Management}, vol.~78, pp. 711--719, 2014.

\bibitem{lin1996learning}
T.~Lin, B.~G. Horne, P.~Tino, and C.~L. Giles, ``Learning long-term
  dependencies in {NARX} recurrent neural networks,'' \emph{IEEE Transactions
  on Neural Networks}, vol.~7, no.~6, pp. 1329--1338, 1996.

\bibitem{andalib2009multi}
A.~Andalib and F.~Atry, ``Multi-step ahead forecasts for electricity prices
  using {NARX}: a new approach, a critical analysis of one-step ahead
  forecasts,'' \emph{Energy Conversion and Management}, vol.~50, no.~3, pp.
  739--747, 2009.

\bibitem{chaabane2014hybrid}
N.~Cha{\^a}bane, ``A hybrid arfima and neural network model for electricity
  price prediction,'' \emph{International Journal of Electrical Power \& Energy
  Systems}, vol.~55, pp. 187--194, 2014.

\bibitem{qing2018hourly}
X.~Qing and Y.~Niu, ``Hourly day-ahead solar irradiance prediction using
  weather forecasts by {LSTM},'' \emph{Energy}, vol. 148, pp. 461--468, 2018.

\bibitem{polson2020deep}
M.~Polson and V.~Sokolov, ``Deep learning for energy markets,'' \emph{Applied
  Stochastic Models in Business and Industry}, vol.~36, no.~1, pp. 195--209,
  2020.

\bibitem{PJMdataminer}
\BIBentryALTinterwordspacing
PJM, \emph{Data Miner 2}, 2020 (Accessed August 29, 2020). [Online]. Available:
  \url{http://dataminer2.pjm.com/list}
\BIBentrySTDinterwordspacing

\bibitem{weatherAPI}
\BIBentryALTinterwordspacing
T.~W. Channel, \emph{Weather.com API}, 2020 (Accessed August 29, 2020).
  [Online]. Available: \url{http://api.weather.com/}
\BIBentrySTDinterwordspacing

\bibitem{weron2004modeling}
R.~Weron, M.~Bierbrauer, and S.~Tr{\"u}ck, ``Modeling electricity prices: jump
  diffusion and regime switching,'' \emph{Physica A: Statistical Mechanics and
  its Applications}, vol. 336, no. 1-2, pp. 39--48, 2004.

\bibitem{lakshminarayanan2017simple}
B.~Lakshminarayanan, A.~Pritzel, and C.~Blundell, ``Simple and scalable
  predictive uncertainty estimation using deep ensembles,'' in \emph{Advances
  in Neural Information Processing Systems}, 2017, pp. 6402--6413.

\end{thebibliography}
\bibliographystyle{IEEEtran}

\end{document}